\documentclass[letterpaper]{article} 
\usepackage{aaai2027}
\nocopyright
\usepackage[hyphens]{url} 
\usepackage{graphicx} 
\usepackage{natbib} 
\usepackage{caption} 
\usepackage{algorithm}
\usepackage{algorithmic}
\usepackage{amsmath}
\usepackage{multirow}
\usepackage{xcolor}

\usepackage{makecell}
\usepackage{newfloat}
\usepackage{listings}
\DeclareCaptionStyle{ruled}{labelfont=normalfont,labelsep=colon,strut=off}
\floatstyle{ruled}
\newfloat{listing}{tb}{lst}{}
\floatname{listing}{Listing}

\usepackage{booktabs}

\title{EAT: Expert Account Tracker for Efficient MoE Inference}

\author{
    Yuexian Li\textsuperscript{\rm 1}\equalcontrib,
    Yifei Yang\textsuperscript{\rm 2,3}\equalcontrib,
    Zouying Cao\textsuperscript{\rm 2,3,4},
    Hai Zhao\textsuperscript{\rm 2,3}\corresponding
}
\affiliations{
    \textsuperscript{\rm 1}Paris Elite Institute of Technology, Shanghai Jiao Tong University\\
    \textsuperscript{\rm 2}AGI Institute, School of Computer Science, Shanghai Jiao Tong University, Shanghai, China\\
    \textsuperscript{\rm 3}Key Laboratory of Shanghai Education Commission for Intelligent Interaction and Cognitive Engineering, \\ Shanghai Jiao Tong University\\
    \textsuperscript{\rm 4}Ant Group\\
    \{liyuexian, yifeiyang\}@sjtu.edu.cn, zhaohai@cs.sjtu.edu.cn
}

\begin{document}

\maketitle
\begin{abstract}
Mixture-of-Experts (MoE) models have emerged as a revolutionary method to scale Transformer models. However, traditional MoE architecture still suffers from inefficiency since a large number of experts are unnecessarily activated. Existing approaches for reducing the number of activated experts often overlook the historical performance of each expert. In this paper, we propose EAT, a novel method called \textbf{Expert Account Tracker (EAT)}, which utilizes history-awareness metrics and adaptive thresholding to dynamically select the most important experts, thereby reducing the activated expert number while effectively maintaining the model performance. Experiments show that EAT outperforms the existing baseline Top-P method across multiple models and datasets, achieving over 25\% an average reduction compared to the vanilla method in the number of activated experts and performing better token generation speed compared to the baseline. 
Furthermore, the performance of pruned models
can be efficiently recovered via OPD using only 9K data.
Additionally, through ablation studies, we find that excessively reducing the number of activated experts can significantly harm model performance, and the importance of experts varies across layers, with higher-level experts being generally more critical.
\end{abstract}

\section{Introduction}

The Mixture-of-Experts (MoE) architecture~\citep{jacobs1991adaptive} has become a prevalent paradigm for scaling large language models (LLMs) with high computational efficiency. 
It replaces the vanilla feed-forward network (FFN) in Transformer blocks with multiple specialized expert sub-networks and employs a learnable router to sparsely activate token-relevant experts during inference~\citep{shazeer2017outrageously, lepikhin2020gshard, fedus2022switch} . 
Benefiting from sparse activation patterns, MoE models exhibit superior scalability compared to dense Transformers, enabling the deployment of large-capacity models with marginal increases in inference overhead~\citep{ku2024proceedings}.  

Despite the inherent sparsity advantages of MoE routing, existing expert activation strategies suffer from two critical limitations. First, vanilla MoE routing adopts fixed Top-K activation~\citep{lepikhin2020gshard} for all input tokens, regardless of input heterogeneity and individual expert utility. This rigid paradigm triggers redundant expert activation and fails to yield optimal sparsity-efficiency balance~\citep{dutt2025exploiting}; 
Second, existing dynamic routing schemes such as Top-P activation~\citep{huang2024harder} rely solely on instantaneous token-level routing scores for expert selection. They completely discard the long-term historical behavior of experts, making it impossible to distinguish consistently reliable experts from unstable, noise-prone experts. This ignorance leads to suboptimal sparse activation: blindly reducing activated expert numbers easily causes severe performance degradation, while conservative activation limits further inference acceleration.
 
To address these issues, we propose a novel  \textbf{E}xpert \textbf{A}ccount \textbf{T}racker (\textbf{EAT}) strategy for dynamic and robust MoE expert routing. 
Unlike existing methods that rely solely on instantaneous gating signals, EAT integrates long-term expert statistics with real-time routing cues to yield more stable and reliable expert importance estimation. 
Additionally, to mitigate the potential performance degradation brought by expert pruning, we adopt a lightweight Optimized Performance Recovery (OPD) phase to calibrate and restore model capacity via small-scale post-optimization tuning.
We evaluate our EAT strategy on mainstream MoE architectures with multiple inference settings. Experimental results demonstrate that our method consistently outperforms vanilla Top-P MoE baselines, reducing the average number of activated experts by 25\% and substantially accelerating token generation speed with negligible performance loss. 

Our key contributions are summarized as follows.
\begin{itemize}
\item \textbf{Dynamic History-Aware MoE Routing.} We propose EAT, which integrates long-term expert historical characteristics and real-time token routing signals via adaptive thresholding.
\item \textbf{Lightweight Performance Recovery Mechanism.} We adopt the lightweight OPD, which only requires a small-scale training set consisting of 9,000 samples to mitigate performance loss. 

\item \textbf{Comprehensive Analysis and Deployment Guidance.} We conduct systematic ablation studies to reveal the intrinsic interplay among model perplexity, expert activation sparsity, and layer-wise routing properties. 
\end{itemize}

\section{Related Works}

\subsection{Evolution of MoE Architectures}
The MoE paradigm has become a mainstream sparse architectural solution for large language models. The pioneering sparsely-gated MoE design~\citep{shazeer2017outrageously} laid the foundation for modern LLM scaling, which was subsequently industrialized by GShard~\citep{lepikhin2020gshard} and Switch Transformer~\citep{fedus2022switch}. These early MoE systems established the canonical fixed Top-K routing mechanism, activating a predetermined number of experts per token and adopting auxiliary losses to alleviate expert load imbalance. Such designs effectively scale model capacity to trillions of parameters while maintaining constant per-token inference cost.

In recent years, open-source MoE models such as Mixtral~\citep{jiang2024mixtral}, DeepSeek-MoE~\citep{xu2026deepseek} have further advanced practical MoE deployment. These open-weight models are tailored for lightweight, accessible inference on consumer and mid-tier computing hardware. They incorporate refined expert partitioning strategies, partial shared feed-forward sub-networks, flexible Top-K routing configurations, and balanced load regularization losses to simultaneously boost real-world inference throughput and cross-domain generalization performance.

\subsection{MoE Efficiency Optimization Methods} 

Beyond the conventional fixed Top-K routing paradigm, efforts have been devoted to improving the efficiency-performance trade-off of MoE models. 

\textbf{Static parameter compression} permanently alters model structure~\citep{lu2024not}. Typical techniques include Merging Experts and Dropping Experts. Moe-pruner~\citep{xie2024moe} prunes weights based on router weight information and weight magnitudes. Channel Merging~\citep{zhang2025channel}  reduces the number of experts by merging parameters
according to similarity. Sub-MoE~\citep{li2025sub} removes experts with low usage. $\text{MoE-I}^2$~\citep{yang2024moe} adopts a two-stage MoE compression framework to reduce model size and computational cost. DiEP~\citep{bai2026diep} conducts offline differentiable expert pruning. 

\textbf{Inference-stage acceleration} increases activation sparsity without modifying model parameters. DA-MoE~\cite {aghdam2024moe} adjusts per-token activation K via token importance signals without permanently removing experts. CMoE~\citep{pei2025cmoe} reorganizes all dense FFN parameters into a combination of ``shared experts'' and ``routed experts''.  Dynamic MoE ~\citep{huang2024harder} sets a routing-probability threshold to retain only higher‑confidence experts. XMoE~\citep{yang2024xmoe} enables each token to autonomously determine the number of experts to activate by using a predefined probability threshold. Ada-K~\citep{yue2024ada} uses a learnable allocator module and a reinforcement learning framework algorithm to adjust the number of activated experts. Expert Pruning And Skipping~\citep{lu2024not} skips less important experts based on the ratio between top routing weights. However, existing inference acceleration methods purely rely on instantaneous token-level routing scores and neglect long-term expert historical reliability. Such single-source routing supervision yields unstable activation decisions under fluctuating input distributions, limiting the sparsity upper bound and failing to achieve a more robust efficiency-performance trade-off.

\section{Method}

\subsection{Vanilla MOE Architecture}

As depicted in Figure~\ref{fig:method}, the MoE architecture differs fundamentally from traditional dense Transformer~\citep{vaswani2017attention} models via a sparse activation paradigm.
Structurally, MoE layers comprise two core modules: Expert Modules and a lightweight gating router. The Expert Modules consist of independent feed-forward sub-networks that substitute standard FFN layers in vanilla Transformers, while the gating router dynamically computes matching scores between input tokens and all experts to select salient experts and assign corresponding computational weights.

During inference, the router first processes the hidden states from the preceding attention layer to generate raw routing scores. It then selects the top-matching experts based on these scores and normalizes the corresponding routing weights. Finally, the MoE layer aggregates the outputs of the selected experts via weighted summation to produce the final layer representation. This token-wise sparse activation mechanism enables MoE models to scale model capacity efficiently while maintaining a controllable inference computational cost.

\subsection{Comprehensive Expert Scoring}

\begin{figure*}[!tp]
    \centering
    \includegraphics[width=0.92\linewidth]{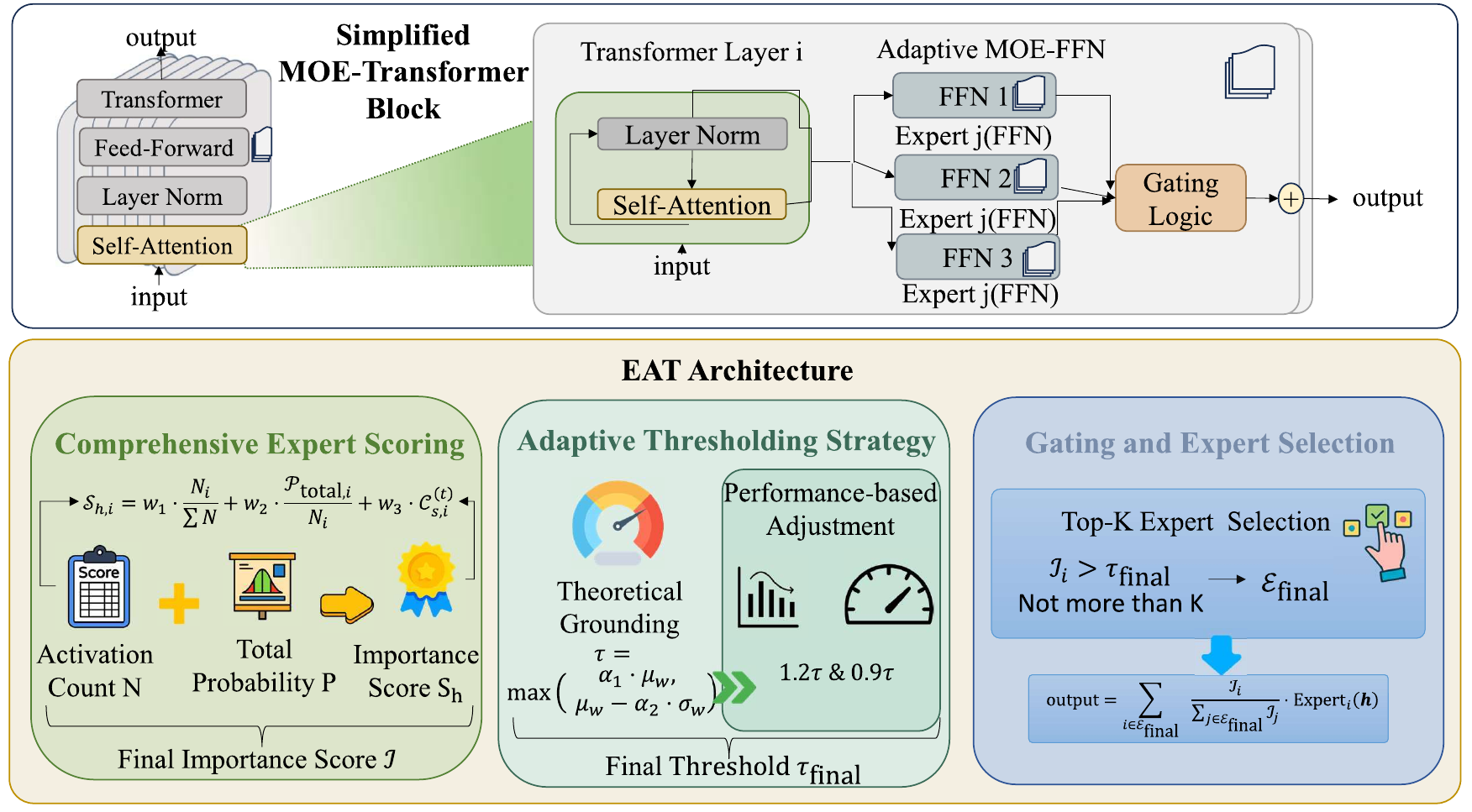}
    \caption{Overview of MoE architecture and our proposed EAT method. EAT selects the most useful experts by continuously tracking their historical contributions and incorporating this information into the routing decision.}
    \label{fig:method}
\end{figure*}

As illustrated in Figure~\ref{fig:method}, EAT dynamically selects experts by integrating a multi-dimensional scoring metric with an adaptive threshold. Unlike vanilla MoE, EAT achieves finer-grained expert sparsification to boost inference efficiency. To address imbalanced expert contributions across layers and balance computational cost and predictive performance, we quantify expert utility from three complementary perspectives as detailed below.

We first design three metrics to calculate the \textbf{history-aware importance score $\mathcal{S}_h$}: 1) \textbf{Activation count $N$}: the total activation frequency of an expert over all inference samples; 
2) \textbf{Total gating probability $\mathcal{P}_{total}$}: the cumulative sum of router-assigned weights collected across all activations of the expert;
3) \textbf{Exponentially smoothed contribution score $\mathcal{C}_s^{(t)}$}: an EMA-based running estimate of the expert’s output magnitude formulated as $\mathcal{C}_s^{(t)} = (1-\alpha) \cdot \mathcal{C}_s^{(t-1)} + \alpha \cdot \|o\|_2$, where
$t$ denotes the current inference timestep, $\alpha \in [0,1]$ is the smoothing factor and $\|o\|_2$ is the L2 norm of the expert output; larger $\mathcal{C}_s^{(t)}$ signifies greater impact on model predictions. Finally, we aggregate these three metrics via weighted summation for expert $i$:
\begin{equation}\label{eq: loss5}
\mathcal{S}_{h,i} = w_1 \cdot \frac{N_i}{\sum N} + w_2 \cdot \frac{\mathcal{P}_{\text{total},i}}{N_i} + w_3 \cdot \mathcal{C}_{s,i}^{(t)},
\end{equation}
where $w_1$, $w_2$, and $w_3$ are weighting hyperparameters, whose specific values are listed in Table~\ref{tab:hyper}.

An expert with a large $\mathcal{S}_{h,i}$ may contribute to the current token when its instantaneous gating weight is low.
To fuse real-time routing signals and long-term expert performance jointly, we construct the \textbf{final importance score}  $\mathcal{I}$ for expert $i$. Let $p_i$ denote the router-assigned selection probability of expert $i$ among all $N$ experts; the unified importance metric is formulated as:
\begin{equation}\label{eq: loss6}
\mathcal{I}_i = (1 - \beta) \cdot p_i + \beta \cdot \mathcal{S}_{h,i},
\end{equation}
where balancing coefficient $\beta$ controls the trade-off between instant routing feedback and historical expert utility, with its detailed value provided in Table~\ref{tab:hyper}.

\subsection{Adaptive thresholding strategy}

After computing expert importance scores $\mathcal{I}$, we dynamically select activated experts via an adaptive threshold $\tau_{\text{final}}$, which is decided by \textbf{theoretical grounding, performance-based adjustment, and boundedness enforcement}.

\paragraph{Theoretical Grounding.} For the expert $i$ of total $N$, the calculation of $\tau$ is decided by \textbf{mean gating weight $\mu_{w}$},  \textbf{standard deviation $\sigma_w$} and \textbf{raw router gating score $g$}:
\begin{equation}
\mu_{w} = \frac{1}{N} \sum_{i=1}^{N} g_i,
\end{equation}
\begin{equation}
    \quad\sigma_{w}= \sqrt{\frac{1}{N} \sum_{i=1}^{N} (g_i - \mu_{w})^2},
\end{equation}
\begin{equation}
 \tau = \max(\alpha_1 \cdot \mu_{w},\; \mu_{w} - \alpha_2 \cdot \sigma_{w}),
 \end{equation}
 where $\alpha_1$ and $\alpha_2$ are two tunable scaling hyperparameters detailed in Table~\ref{tab:hyper}. The term $\alpha_1 \cdot \mu_w$ enforces a minimal activation floor to avoid severe performance degradation under highly scattered routing weights.
 
\paragraph{Performance-based Adjustment.} MoE routing distributions exhibit non-stationary variations due to fluctuating input difficulty. To address this issue, we dynamically calibrate $\tau$ according to real-time model perplexity (PPL). $\text{PPL}_{\text{curr}} - \text{PPL}_{\text{prev}} > 0$ indicates that the current perplexity is higher than before, meaning a drop in performance. To activate more experts and improve performance, the threshold becomes $\tau \cdot (1 + 20\%)$; otherwise, it will be $\tau \cdot (1 - 10\%)$.

\paragraph{Boundedness Enforcement.} To ensure stability and prevent extreme activation or sparsity, the final threshold is constrained to a predefined range, $\tau_{\text{final}} \in [\tau_{\text{min}}, \tau_{\text{max}}]$. In our implementation, $\tau_{\text{min}}$ and $\tau_{\text{max}}$ are set as 10\% and 90\% of the routing weight distribution, respectively, ensuring a minimum level of expert activation.
Finally, we adopt the exponential smoothing transition to avoid abrupt changes in the threshold caused by performance fluctuations: $\tau_{\text{final}} = \kappa_1 \cdot \tau_{\text{PPL}} + \kappa_2 \cdot \tau$
where $\kappa_1$ and $\kappa_2$ are two hyperparameters representing weights.

\subsection{Expert Activation Mechanism}
During inference, we vary the number of selected experts by comparing $\mathcal{I}_i$ against  $\tau_{\text{final}}$. 
The candidate expert set $\mathcal{E}$ is defined as: $\mathcal{E} = \{i \mid \mathcal{I}_i > \tau_{\text{final}}\}$. 
After being compared with the $K$ of Top-$K$: 
\begin{equation}
\mathcal{E}_{\text{final}} = 
\begin{cases} 
\{\arg\max_{i} \mathcal{I}_i\} & |\mathcal{E}| = 0, \\
\mathcal{E} & 1 \leq |\mathcal{E}| \leq K, \\
\left\{ i \in \mathcal{E} \,\big|\, |\{j \in \mathcal{E} \mid \mathcal{I}_j > \mathcal{I}_i\}| < K \right\} & |\mathcal{E}| > K,
\end{cases}
\end{equation}

After expert selection, we perform weight renormalization and define the binary activation mask $m_i$: 
\begin{equation}
    g_{norm,i} = \frac{g_{i} \cdot m_{i}}{\sum_{i=1}^E g_{i} \cdot m_{i} + \epsilon}\;,
\end{equation}
\begin{equation}
    \quad m_i=\begin{cases}1&
        $i$ \in \mathcal{E}_{\text{final}},\\0&\text{otherwise,}
    \end{cases},
\end{equation}
where $\epsilon=10^{-8}$ is a small constant to avoid division by zero. Only activated experts with $m_i=1$ participate in weight normalization and feature fusion. Finally, the MoE layer output is computed via weighted summation over activated experts using the normalized gating weights. This adaptive selection and fusion pipeline enables our EAT-MoE to dynamically accommodate diverse input distributions, achieving an optimal trade-off between inference efficiency and modeling performance.

\subsection{Recovery with OPD}

We integrate OPD into the EAT routing pipeline to mitigate performance degradation induced by sparse expert selection. During each forward pass, we compute and normalize token-averaged routing probabilities to construct an entropy-based regularization loss, which is cached and incorporated into the overall training objective following. As a universal enhancement applicable to all MoE architectures, OPD additionally reweights raw routing probabilities prior to Top-K expert selection by scaling each score and renormalizing the resulting distribution. The OPD balancing coefficient $\lambda_{\text{opd}}$ disables regularization when set to zero by default, while we assign a value of $0.8$ in core experiments to strengthen uniform expert activation. We maintain consistent training and generation configurations across all evaluated models, including a learning rate of $5\times10^{-5}$, single-step gradient accumulation, a maximum prompt length of 256, a maximum output length of 64, sampling temperature 0.8, with only router weights updated throughout fine-tuning. The details of selection of steps and dataset size are in Figure~\ref{fig:number_dataset} and Figure~\ref{fig:number_step}.

\section{Experiments}
\subsection{Experimental Setup}
To thoroughly assess the effectiveness of the proposed EAT method, we conduct extensive experiments on popular MoE LLMs and a comprehensive benchmark suite.

\paragraph{Models}
Our experiment utilizes three popular MoE LLMs: \textbf{Mixtral-8x7B-v0.1}~\citep{jiang2024mixtral}, \textbf{Phi-3.5-MoE-instruct}~\citep{abdin2024phi} and \textbf{Qwen3-30B-A3B}~\citep{yang2025qwen3}. These models are chosen for their MoE architectures and widespread recognition. We ensure that our findings on efficiency and performance gains are robust and generalizable across different scales and application types. All experiments are performed on 8 NVIDIA A100 GPUs, providing the necessary computational capacity to handle these large-scale models.
\paragraph{Benchmarks}
To comprehensively evaluate the model performance, we employ the OpenCompass evaluation framework~\citep{2023opencompass}. This framework allows us to categorize our evaluations into five key dimensions:  \textbf{Reasoning, Language, Knowledge, Examination, and Understanding}. To ensure a broad and representative evaluation, we select specific benchmarks within each category. Reasoning: HellaSwag (HeSw)~\citep{zellers2019hellaswag} and PIQA~\citep{bisk2019piqa}. Language: CHID~\citep{zheng-etal-2019-chid} and WSC~\citep{levesque2012winograd}. Knowledge: BoolQ~\citep{clark2019boolq}. Examination: MMLU~\citep{hendryckstest2021} and CMMLU~\citep{li2023cmmlu}. Understanding: XSum~\citep{narayan2018don}.

All evaluations were executed using OpenCompass's official scripts, employing two primary evaluation models: \textbf{Perplexity (PPL) and Generation (GEN)}. Specifically, we used the GEN mode for CHID, XSum, and WSC, while the PPL mode was applied to BoolQ, HeSw, PIQA, MMLU, and CMMLU. The final accuracy scores for each benchmark are standardized by OpenCompass, with a higher score indicating superior performance. 
\paragraph{Baselines}
To provide a meaningful comparison, we select the \textbf{Top-$P$} method~\citep{huang2024harder} as our baseline. As both Top-$K$ and Top-$P$ are widely recognized MoE LLM methods, choosing the Top-$P$ method allows us to directly compare our EAT's performance with a method that dynamically adjusts the number of activated experts. To ensure that the number of activated experts is similar to that of our EAT method, we set a constant $P$=0.6 for our Mixtral baseline, $P$=0.2 for Phi baseline and $P$=0.4 for Qwen baseline. 

\paragraph{Hyperparameters}
We tabulate the hyperparameters in our experiments in Table~\ref{tab:hyper}:

\begin{table}[t]
\centering
\small
\setlength\tabcolsep{4pt}
\renewcommand{\arraystretch}{1.1}
\begin{tabular}{@{}l l p{3.4cm}@{}}
\toprule
Hyperparameter & Value & Description \\
\midrule
$\alpha$ & 0.95 & Smoothing factor for $\mathcal{C}_s^{(t)}$.\\
$w_1,w_2,w_3$ & 0.3, 0.3, 0.4 & Weights to compute $\mathcal{S}_{h,i}$. \\
$\beta$ & 0.6 & Balance factor for final importance $\mathcal{I}_i$. \\
$\alpha_1,\alpha_2$ & 0.5, 1.0 & Scaling coefficients for threshold $\tau$. \\
$\kappa_1,\kappa_2$ & 0.7, 0.3 & Weights for threshold adjustment. \\
$K$ (Mixtral/Phi) & 2 & Max activated experts per token. \\
$K$ (Qwen) & 8 & Max activated experts per token. \\
\bottomrule
\end{tabular}
\caption{Key hyperparameters of the proposed EAT routing strategy.}
\label{tab:hyper}
\end{table}

\begin{table*}[!t]
    \setlength\tabcolsep{5pt} 
    \centering
    \renewcommand{\arraystretch}{1.1}
    \resizebox{0.97\linewidth}{!}{
    \begin{tabular}{cccccccccc}
    \toprule[0.5pt]
    \multirow{2}{*}{\textbf{LLM}} & \multirow{2}{*}{\textbf{Method}} 
    & \multicolumn{2}{c}{\textbf{Reasoning}} 
    & \multicolumn{2}{c}{\textbf{Language}} 
    & \multicolumn{1}{c}{\textbf{Know.}} 
    & \multicolumn{2}{c}{\textbf{Examination}} 
    & \multicolumn{1}{c}{\textbf{Und.}} \\
    & & HeSw & PIQA & CHID & WSC & BoolQ & MMLU & CMMLU & XSum \\
    \midrule
    \multirow{4}{*}{\textbf{\makecell[c]{Mixtral \\ -8x7B \\ -v0.1}}}
    & Vanilla & 77.11 & 81.07 & 37.51 & 61.54 & 69.11 & 71.67 & 53.11 & 9.19 \\
    & Top-P   & 75.72 & 79.71 & 32.85 & 60.58 & 66.15 & 65.38 & 46.74 & 8.70 \\
    \cmidrule{2-10}
    & EAT     & 76.79 & 80.30 & 33.97 & 63.46 & 68.40 & 70.37 & 51.04 & 9.08 \\
    & EAT + OPD & 77.05 & 80.96 & 35.42 & 64.83 & 69.25 & 71.59 & 52.66 & 9.10 \\
    \midrule
    \multirow{4}{*}{\textbf{\makecell[c]{Phi \\ -3.5-MoE \\ -instruct}}}
    & Vanilla & 75.17 & 80.20 & 66.75 & 68.27 & 75.32 & 76.64 & 61.03 & 14.68 \\
    & Top-P   & 68.98 & 78.67 & 60.99 & 71.15 & 69.36 & 74.05 & 55.26 & 11.32 \\
    \cmidrule{2-10}
    & EAT     & 69.00 & 79.87 & 61.00 & 67.31 & 76.42 & 75.00 & 57.46 & 12.46 \\
    & EAT + OPD & 72.19 & 79.93 & 65.70 & 67.88 & 77.60 & 74.89 & 59.62 & 13.67 \\
    \midrule
    \multirow{4}{*}{\textbf{\makecell[c]{Qwen3 \\ -30B-A3B}}}
    & Vanilla & 70.93 & 80.63 & 44.81 & 78.86 & 79.02 & 80.22 & 81.73 & 12.21 \\
    & Top-P   & 68.50 & 76.61 & 32.67 & 78.85 & 77.16 & 77.85 & 78.94 & 9.93 \\
    \cmidrule{2-10}
    & EAT     & 69.52 & 79.82 & 50.61 & 72.12 & 74.74 & 73.96 & 78.89 & 8.90 \\
    & EAT + OPD & 70.56 & 78.89 & 67.46 & 74.20 & 78.26 & 79.65 & 80.45 & 18.81 \\
    \bottomrule
    \end{tabular}
    }
    \caption{Main experimental results evaluated on the OpenCompass Platform. \textbf{Know.} denotes Knowledge benchmarks and \textbf{Und.} denotes Understanding benchmark.}
    \label{tab:main_res}
\end{table*}

\subsection{Main Results}
\label{sec:results}

The main results are summarized in Table~\ref{tab:main_res}, Table~\ref{tab:main_resex}, and Table~\ref{tab:tokenspeed}, which clearly demonstrate that our EAT model effectively balances computational efficiency with model performance in various tasks and models.

\begin{table*}[!t]
    \setlength\tabcolsep{5pt} 
    \centering
    \renewcommand{\arraystretch}{1.1}
    \resizebox{0.97\linewidth}{!}{
    \begin{tabular}{cc c c c c c c c c}
    \toprule
    \multirow{2}{*}{\textbf{LLM}} & \multirow{2}{*}{\textbf{Method}} 
    & \multicolumn{2}{c}{\textbf{Reasoning}} 
    & \multicolumn{2}{c}{\textbf{Language}} 
    & \multicolumn{1}{c}{\textbf{Know.}} 
    & \multicolumn{2}{c}{\textbf{Examination}} 
    & \multicolumn{1}{c}{\textbf{Und.}} \\
    & & HeSw & PIQA & CHID & WSC & BoolQ & MMLU & CMMLU & XSum \\
    \midrule
    \multirow{4}{*}{\textbf{\makecell[c]{Mixtral \\ -8x7B \\ -v0.1}}} 
    & Vanilla & 2.00  & 2.00  & 2.00  & 2.00  & 2.00  & 2.00  & 2.00  & 2.00 \\
    & Top-P   & 1.50  & 1.44  & 1.75  & 1.52  & 1.46  & 1.51  & 1.74  & 1.51 \\
    \cmidrule{2-10}
    & EAT     & 1.47  & 1.46  & 1.44 & 1.50 & 1.45 & 1.47 & 1.48 & 1.47 \\
    & EAT + OPD & 1.50 & 1.47 & 1.46 & 1.50 & 1.45 & 1.69 & 1.57 & 1.71 \\
    \midrule
    \multirow{4}{*}{\textbf{\makecell[c]{Phi \\ -3.5-MoE \\ -instruct}}}
    & Vanilla & 2.00 & 2.00 & 2.00 & 2.00 & 2.00 & 2.00 & 2.00 & 2.00 \\
    & Top-P   & 1.32 & 1.67 & 1.74 & 1.72 & 1.71 & 1.76 & 1.75 & 1.73 \\
    \cmidrule{2-10}
    & EAT     & 1.36 & 1.34 & 1.50 & 1.35 & 1.70 & 1.43 & 1.44 & 1.36 \\
    & EAT + OPD & 1.47 & 1.34 & 1.68 & 1.35 & 1.73 & 1.61 & 1.44 & 1.37 \\
    \midrule
    \multirow{4}{*}{\textbf{\makecell[c]{Qwen3 \\ -30B-A3B}}}
    & Vanilla & 8.00  & 8.00  & 8.00  & 8.00  & 8.00  & 8.00  & 8.00  & 8.00 \\
    & Top-P   & 6.38  & 6.65  & 6.46  & 5.98  & 6.48  & 6.75  & 6.64  & 6.30 \\
    \cmidrule{2-10}
    & EAT     & 6.39  & 5.90  & 6.07  & 5.93  & 6.15  & 5.50  & 5.67  & 6.28 \\
    & EAT + OPD & 6.59 & 6.42 & 6.07 & 5.93 & 6.26 & 6.10 & 5.63 & 6.63 \\
    \bottomrule
    \end{tabular}
    }
    \caption{Comparison of the average number of activated experts under different expert routing strategies.}
    \label{tab:main_resex}
\end{table*}

\begin{table}[!tp]
    \setlength\tabcolsep{4pt}
    \small
    \centering
    \renewcommand{\arraystretch}{1.3}
    \begin{tabular}{@{}l c c c c c c @{}} 
    \toprule
    Length & \multicolumn{3}{c}{\textbf{Mixtral-8x7B}} & \multicolumn{3}{c}{\textbf{Phi-3.5-MoE}} \\
    \cmidrule(r){2-4} \cmidrule(l){5-7}
          & Vanilla & Top-P & EAT & Vanilla & Top-P & EAT \\
    \midrule
    2048+256 & 0.5214 & 0.5100 & \textbf{0.5196} & 1.1319 & 0.9479 & \textbf{1.1600} \\
    1024+128 & 0.3931 & 0.3739 & 0.3864 & 0.9295 & 0.9073 & 0.9158 \\
    512+32   & 0.3814 & 0.1596 & 0.3671 & 0.8060 & 0.8719 & \textbf{1.1421} \\
    \bottomrule
    \end{tabular}
    \caption{Average token generation speed (token/sec).}
    \label{tab:tokenspeed}
\end{table}

\textbf{EAT Outperforms Baselines in Performance and Efficiency}. As shown in Table~\ref{tab:main_res}, EAT method generally outperforms the Top-P baselines across a range of benchmarks and evaluation modes, consistently achieving higher scores on reasoning, language, and knowledge tasks; while exhibiting a slight decrease in score compared to the vanilla model, it activates fewer experts, which is a key finding indicating that EAT can achieve performance similar to the vanilla model with much greater computational efficiency. 
When compared directly to the Top-P method, which activates a similar number of experts, our EAT method shows great performance across all datasets evaluated. For Mixtral-8x7B-v0.1, EAT improves core metrics including WSC from 60.58 for Top-P to 63.46, and BoolQ from 66.15 for Top-P to 68.40 without notable regression on reasoning tasks like HellaSwag and PIQA. On Phi-3.5-MoE-instruct, EAT brings clear gains on knowledge benchmarks: BoolQ rises from 69.36 for Top-P to 76.42, while reasoning performance on PIQA remains nearly intact. For Qwen3-30B-A3B, EAT greatly lifts language comprehension metrics such as CHID from 32.67 for Top-P method to 50.61. This highlights a major advantage of EAT's adaptive approach, which considers an expert's historical performance rather than just relying on its current routing score, ensuring a more effective and higher-quality expert activation.

\textbf{EAT Provides a Significant Reduction in Activated Experts}. Table~\ref{tab:main_resex} clearly illustrates that EAT method obviously activates fewer experts than the vanilla models across all categories, achieving a reduction of nearly 25\%. On the Mixtral-8x7B-v0.1 model, the vanilla method activates 2 experts for reasoning tasks, while EAT activates only 1.46 on average. For the Phi-3.5-MoE-instruct model, the number of activated experts for the same reasoning tasks drops from 2 with the vanilla method to 1.34 with EAT at best. For Qwen3-30B-A3B which adopts a vanilla setting with 8 fixed activated experts, under reasoning benchmarks, EAT shrinks the average activated experts to 6.39 on HeSw and 5.90 on PIQA, delivering a great reduction compared with the vanilla model. This large reduction in the number of activated experts is a direct result of EAT's dynamic and history-aware selection strategy, which significantly lowers the computational overhead per token.

\textbf{EAT Achieves Faster Token Generation Speed}.
We further evaluate token generation speed, with results summarized in Table~\ref{tab:tokenspeed}. 
EAT delivers generation speeds on par with vanilla MoE and outperforms standard Top-P routing baselines across all evaluated settings. When deploying Mixtral-8x7B with a 1024-token input context and a maximum output length of 128 tokens, EAT yields a generation rate of 0.3864 tokens per second. This performance marginally lags vanilla MoE at 0.3931 tokens per second, yet clearly exceeds Top-P routing which attains only 0.3739 tokens per second. For short-sequence inference on Phi-3.5-MoE-instruct with a 512-token input context and a maximum output length of 32 tokens, EAT achieves a throughput of 1.1421 tokens per second. This rate surpasses both vanilla MoE at 0.8060 tokens per second and Top-P routing at 0.8719 tokens per second. These consistent observations verify that the inference optimizations embedded within EAT yield substantial practical efficiency gains.

\textbf{OPD Mitigates The Performance Loss}.
The proposed OPD lightweight optimization effectively compensates for the performance degradation caused by aggressive expert sparsification. As shown in Table~\ref{tab:main_res}, for Qwen3-30B-A3B model, the EAT + OPD paradigm obtains great gains on key benchmarks, lifting the CHID score from 50.61 to 67.46, and the XSum metric from 8.90 to 18.81. For Mixtral-8x7B-v0.1, equipping EAT with OPD yields steady improvements across most benchmarks, notably boosting the CHID score from 33.97 to 35.42 while maintaining the low average expert activation count of EAT. For Phi-3.5-MoE-instruct, the EAT+OPD combination delivers visible gains on knowledge and comprehension tasks: the BoolQ metric rises from 76.42 to 77.60, and CMMLU performance is elevated from 57.46 to 59.62. Meanwhile, the expert activation sparsity advantage of EAT is preserved, without sacrificing the inference acceleration capability. These results demonstrate that OPD serves as an efficient and pluggable performance compensation module, which successfully resolves the efficiency-performance trade-off bottleneck of dynamic MoE routing and enables EAT to achieve sparse expert activation and competitive model performance simultaneously.

\section{More Analysis}
\label{sec:analysis}

\begin{figure}[!tp]
    \centering
    \includegraphics[width=0.95\columnwidth]{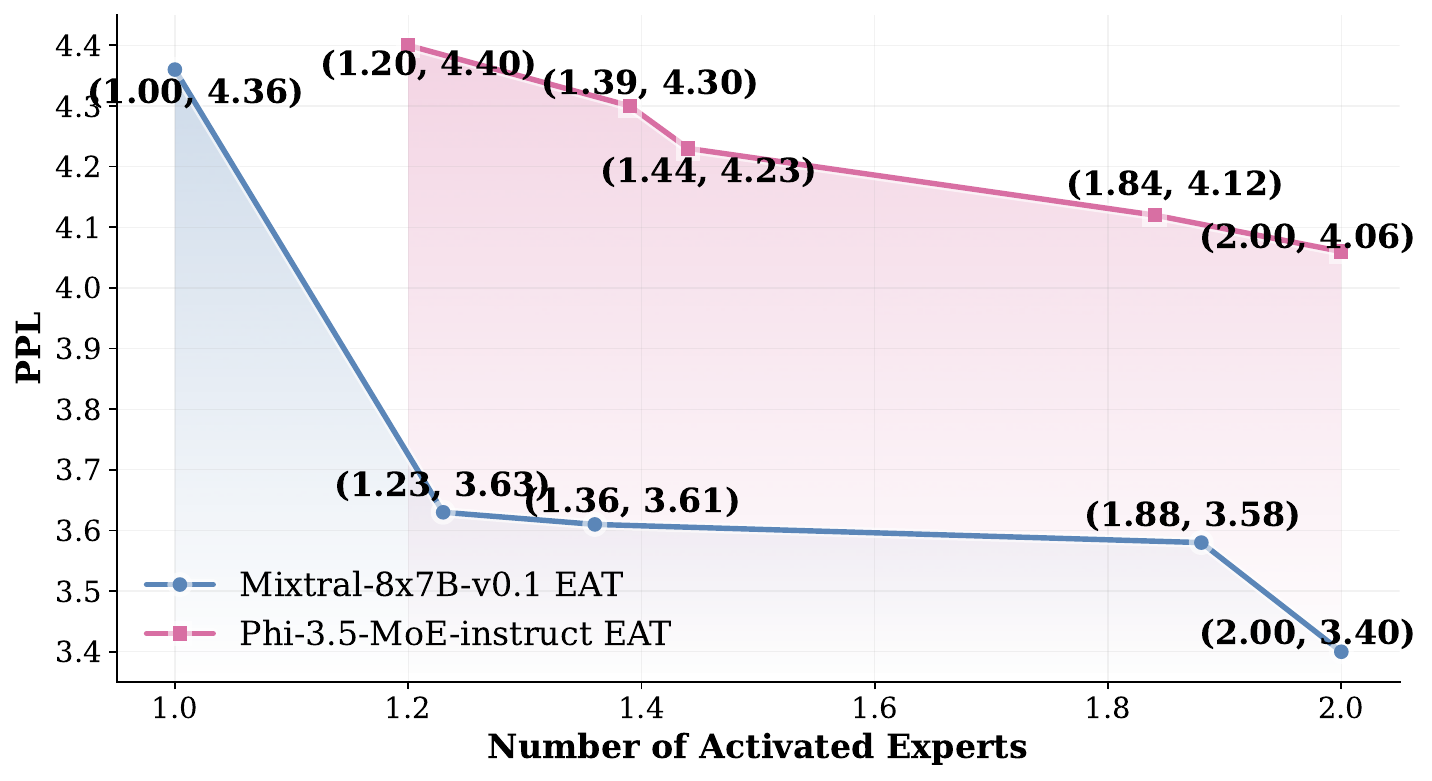}
    \caption{Curve of the Relationship Between PPL and Number of Activated Experts.}
    \label{fig:eat}
\end{figure}

\begin{figure}[!tp]
    \centering
    \includegraphics[width=0.95\columnwidth]{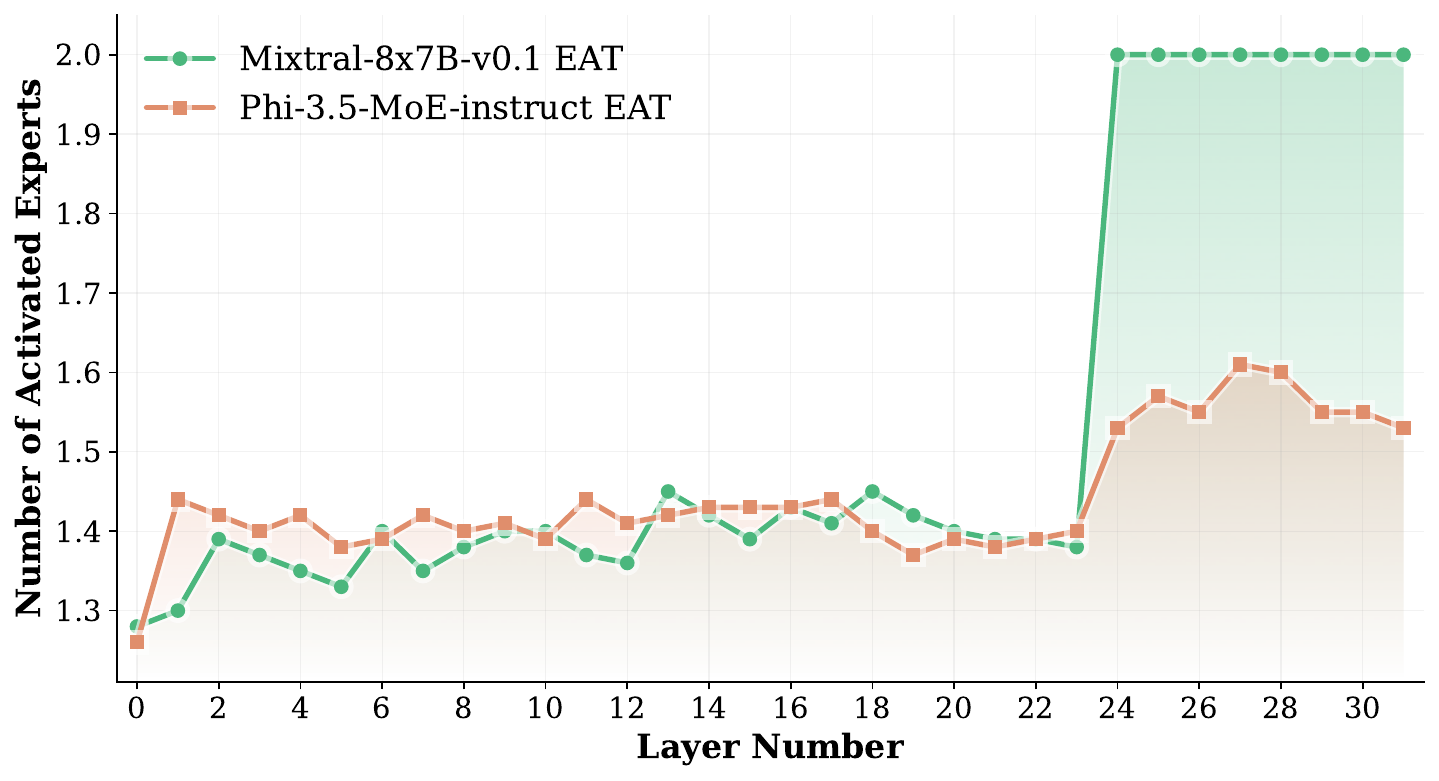}
    \caption{Curve of the Relationship Between layer number and Number of Activated Experts.}
    \label{fig:eat2}
\end{figure}

\subsection{The Impact of the Number of Activated Experts}
We analyze the relationship between the number of activated experts and model performance, measured by \textbf{PPL}. As shown in Figure~\ref{fig:eat}, Curves plotted over Mixtral and Phi reveal a clear negative correlation between PPL and the average number of activated experts: PPL declines monotonically as more experts are activated. The model achieves stable, favorable PPL values at approximately 1.36 activated experts per token, with marginal performance fluctuations around this operating point.

Our analysis further yields a critical observation: aggressive sparsification that drastically cuts activated experts severely degrades model performance. For Mixtral-8x7B-v0.1, restricting each token to only one activated expert pushes PPL sharply up to 4.3, a notable deterioration compared to the optimal activation regime. This result demonstrates that expert sparsity cannot be arbitrarily maximized, and a lower bound on expert activation is essential to retain modeling capacity. Consistent with the above trend, Figure~\ref{fig:eat2} shows that the PPL curve plateaus at an average activation count of 1.36 experts per token, verifying this threshold as a robust point for balanced efficiency and performance.

\subsection{The importance of Experts Across Layers}
Our study also investigates the sparsity and importance of experts across different layers of the MoE models and results in Figure~\ref{fig:eat2}
reveal a crucial architectural insight: the importance and optimal activation of experts are not uniform across layers. 

Specifically, we find that experts in lower layers are less critical than those in higher layers. The lower layers here refer to layers 0 to 23, while the higher layers cover layers 24 to 30. This finding aligns with the hierarchical design of Transformers. In such a design, lower layers typically learn more general features, whereas higher layers are tasked with processing complex information. This characteristic enables us to adopt an aggressive sparsity strategy in lower layers, where we only need to maintain 1.3 to 1.5 activated experts. For higher layers, we can fully activate them. This means activating all experts or nearly all Top-K experts, and such full activation will not lead to significant performance loss. 
This insight is valuable for future model design and training, suggesting a potential for more selective and layered sparsity strategies. Instead of applying the vanilla approach, a dynamic, layer-wise routing mechanism can significantly enhance computational efficiency by activating only the necessary number of experts of the model.

\subsection{The number of steps and datasets of OPD}
\label{sec:opd}

\begin{figure}[!tp]
    \centering
    \includegraphics[width=0.95\columnwidth]{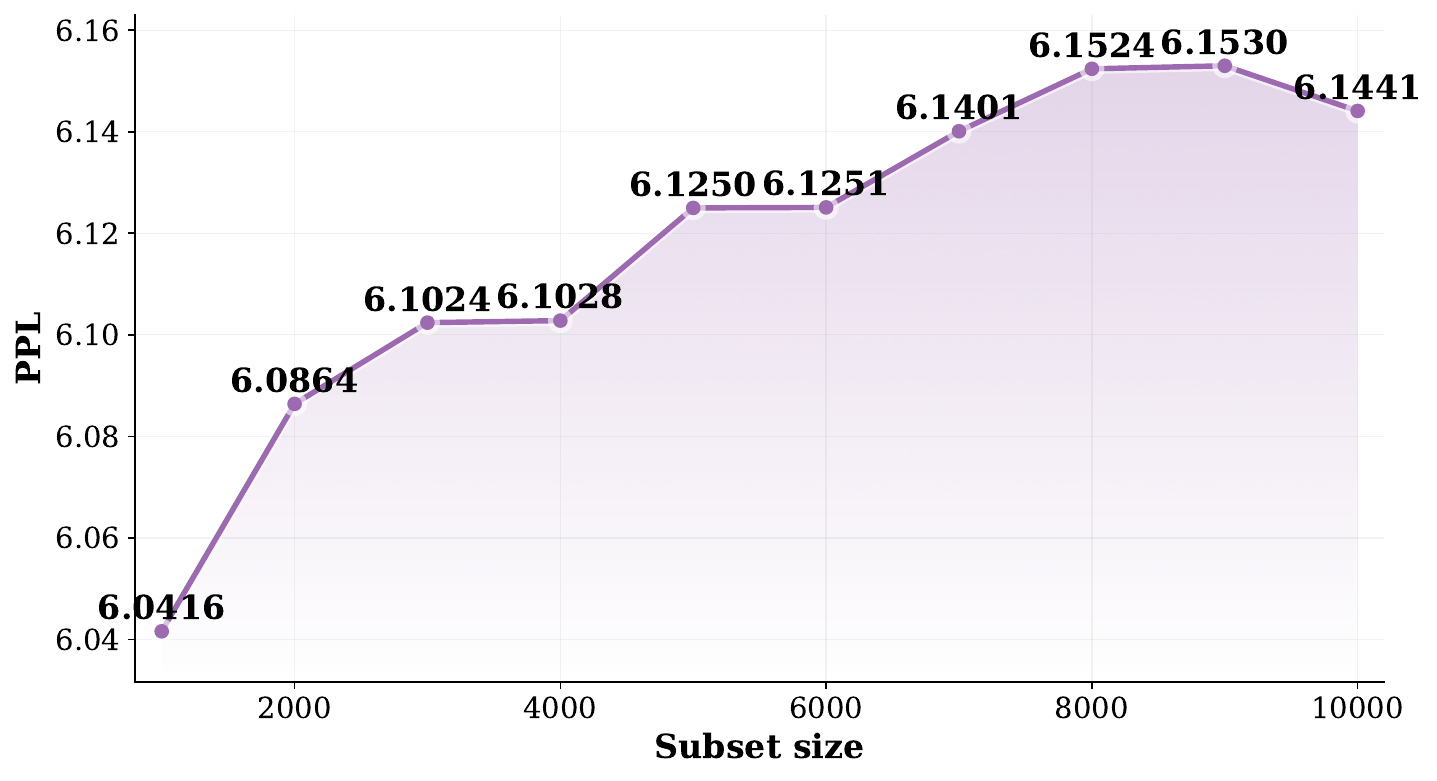}
    \caption{Curve of PPL on OPD with different subset size on Qwen30B on PIQA dataset.}
    \label{fig:number_dataset}
\end{figure}

\begin{figure}[!tp]
    \centering
    \includegraphics[width=0.95\columnwidth]{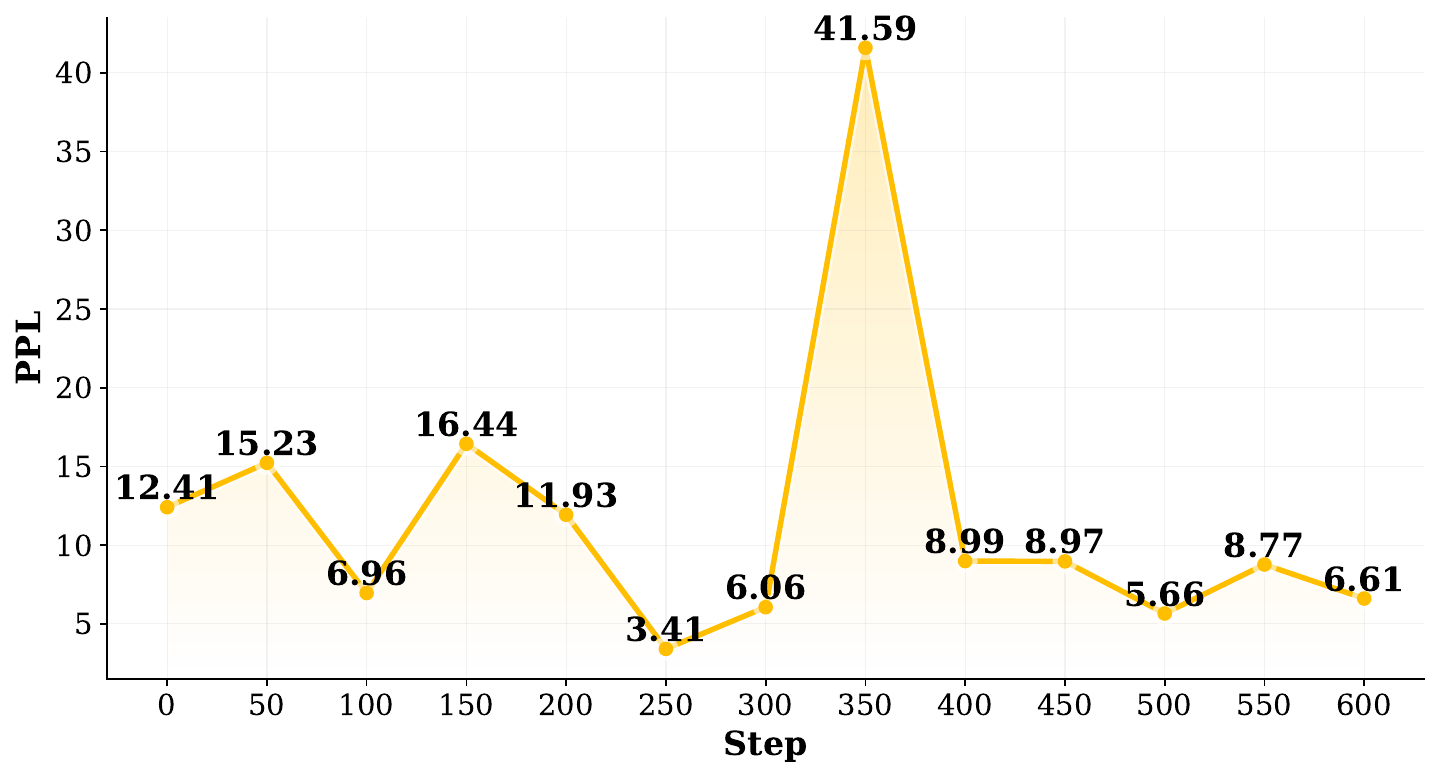}
    \caption{Curve of PPL on OPD with different steps on Qwen30B on BOOLQ dataset.}
    \label{fig:number_step}
\end{figure}

We conduct ablation studies to select optimal hyperparameters for OPD, including tuning steps and training data scale. As illustrated in Figure~\ref{fig:number_dataset}, the model achieves peak performance with a dataset of 9,000 samples. Meanwhile, Figure~\ref{fig:number_step} shows the optimal tuning step count is 250. Excessively large datasets or excessive tuning steps instead degrade the final model performance. To eliminate domain distribution mismatch that may introduce additional performance bias, we construct the OPD datasets identical task categories and data distribution as the test benchmarks adopted throughout our evaluation pipeline. For the reasoning one, we use AX-b~\citep{wang2019superglue}, AX-g~\citep{wang2019superglue}, CB~\citep{de2019commitmentbank}, COPA~\citep{roemmele2011choice}, MultiRC~\citep{khashabi2018looking}, RTE~\citep{wang2018glue}, ReCoRD~\citep{zhang2018record}, WiC~\citep{pilehvar2018wic} and GSM8K~\citep{cobbe2021training}. For the knowledge one, we use commonsenseqa~\citep{talmor2019commonsenseqa}, triviaqa~\citep{joshi2017triviaqa}, nq-open~\citep{kwiatkowski2019natural}, arc-challenge-dev~\citep{clark2018think}, arc-easy-dev~\citep{clark2018think}, winogrande-train-l~\citep{sakaguchi2020winogrande}, gpqa-diamond~\citep{rein2023gpqa}, simpleqa~\citep{wei2024measuring} and TyDiQA~\citep{clark2020tydi}.

\section{Conclusion}
In this paper, we introduce EAT, a lightweight inference optimization framework for sparse MoE models. Unlike conventional MoE routing strategies that rely solely on instantaneous gating scores and neglect long-term expert behavioral characteristics, EAT incorporates a history-aware expert scoring mechanism and a statistically grounded adaptive thresholding strategy to achieve fine-grained, stable expert selection. And we further adopt OPD with a limited sample dataset to efficiently compensate for precision loss. Extensive experiments across diverse models and datasets demonstrate that EAT consistently outperforms existing baseline methods by reducing redundant expert activations while preserving competitive performance. Our ablation further validate the layer-varying expert importance and verify the necessity of bounded expert sparsification. In summary, EAT represents a robust and flexible strategy that realizes an optimal balance between computational efficiency and model accuracy.

\appendix

\bibliography{aaai2027}

\end{document}